\documentclass[
 two-column
]{ceurart}

\usepackage{listings}
\usepackage{amsmath}
\usepackage{url}
\begin{document}

\copyrightyear{2022}
\copyrightclause{Copyright for this paper by its authors.
  Use permitted under Creative Commons License Attribution 4.0
  International (CC BY 4.0).}

\conference{Forum for Information Retrieval Evaluation, December 12-15, 2024, India}

\title{RetrieveGPT: Merging Prompts and Mathematical Models for Enhanced Code-Mixed Information Retrieval}

%\tnotetext[1]{You can use this document as the template for preparing your
%  publication. We recommend using the latest version of the ceurart style.}

\author[1]{Aniket Deroy}[%
orcid=0000-0001-7190-5040,
email=roydanik18@kgpian.iitkgp.ac.in,
%url=https://yamadharma.github.io/,
]
\cormark[1]
\fnmark[1]
\address[1]{IIT Kharagpur,
  Kharagpur, India}
%\address[2]{}%Joint Institute for Nuclear Research,
  %6 Joliot-Curie, Dubna, Moscow region, 141980, Russian Federation}

\author[1]{Subhankar Maity}[%
orcid=0009-0001-1358-9534,
email=subhankar.ai@kgpian.iitkgp.ac.in,
%url=https://kmitd.github.io/ilaria/,
]
%\address[3]{Vrije Universiteit Amsterdam, De Boelelaan 1105, 1081 HV Amsterdam, The Netherlands}

\cortext[1]{Corresponding author.}
%\fntext[1]{These authors contributed equally.}

\begin{abstract}
Code-mixing, the integration of lexical and grammatical elements from multiple languages within a single sentence, is a widespread linguistic phenomenon, particularly prevalent in multilingual societies. In India, social media users frequently engage in code-mixed conversations using the Roman script, especially among migrant communities who form online groups to share relevant local information. This paper focuses on the challenges of extracting relevant information from code-mixed conversations, specifically within Roman transliterated Bengali mixed with English.
%During the COVID-19 pandemic, the need for reliable information became crucial for migrant communities in India. For example, Bengali speakers in cities like Delhi and Bangalore formed online groups on platforms such as Facebook and WhatsApp to seek and share advice on local issues, including navigating frequently changing government guidelines. These discussions, characterized by informal and colloquial language transliterated into Roman script, pose significant challenges for identifying relevant answers due to the lack of standardization in the text.
This study presents a novel approach to address these challenges by developing a mechanism to automatically identify the most relevant answers from code-mixed conversations. We have experimented with a dataset comprising of queries and documents from Facebook, and Query Relevance files (QRels) to aid in this task. Our results demonstrate the effectiveness of our approach in extracting pertinent information from complex, code-mixed digital conversations, contributing to the broader field of natural language processing in multilingual and informal text environments. We use GPT-3.5 Turbo via prompting alongwith using the sequential nature of relevant documents to frame a mathematical model which helps to detect relevant documents corresponding to a query.

\end{abstract}


\maketitle

\section{Introduction}

Code-mixing, where elements from multiple languages are blended within a single sentence, is a natural and widespread phenomenon in multilingual societies \cite{cm1, cm2}. Code mixing is a global phenomenon where speakers often switch between languages depending on context, audience, and medium of communication \cite{cm3}. With the rapid rise of online social networking, this practice has become increasingly common in digital conversations, where users frequently combine their native languages with others, often using foreign scripts \cite{cm4}.

One notable trend in India is the use of the Roman script to communicate in native languages on social media platforms \cite{cm5}. This practice is especially common among migrant communities who form online groups to share information and experiences relevant to their unique circumstances \cite{cm6}. For instance, Bengali speakers from West Bengal who have migrated to urban centers like Delhi or Bangalore often establish groups such as "Bengali in Delhi" on platforms like Facebook and WhatsApp. These groups serve as vital hubs for exchanging advice on a wide range of local issues, from housing and employment to navigating new social environments.

The COVID-19 pandemic highlighted the importance of these online communities as critical sources of information \cite{cm7}. During this period, these groups became essential for sharing experiences, seeking support, and keeping up with the frequently changing government guidelines. However, the informal and often colloquial nature of the language used in these code-mixed conversations, typically transliterated into Roman script, presents significant challenges for information retrieval. The lack of standardization, combined with the blending of languages, makes it difficult to identify and extract relevant answers, especially for those who might seek similar information at a later time \cite{cm8}.

This paper addresses the challenge of extracting relevant information from code-mixed digital conversations, with a specific focus on Roman transliterated Bengali mixed with English. While code-mixing is a well-recognized phenomenon in natural language processing (NLP), the unique characteristics of transliterated text—such as variations in spelling, grammar, and syntax—complicate the task of effective information retrieval \cite{cm9}. To tackle this issue, we have developed a mechanism that identifies the most relevant answers from these complex, multilingual discussions.

We begin experimenting with a dataset of code-mixed conversations collected from Facebook, which has been carefully annotated to reflect query relevance (QRels). This dataset forms the basis of our study and is crucial for evaluating the effectiveness of our approach.% The remainder of this paper is organized as follows: we first review related work in the field of code-mixing and information retrieval, then detail our methodology for retrieving relevant documents. Following this, we present our findings, demonstrating how our approach can improve the identification of relevant information in code-mixed conversations. Finally, we discuss the implications of our research for NLP in multilingual contexts and outline potential future directions for this work.

We leverage GPT-3.5 Turbo \cite{cm24} by employing carefully designed prompts that guide the model to evaluate the relevance of documents with respect to a given query. This involves not only the semantic understanding capabilities of GPT-3.5 Turbo but also the strategic use of the sequential nature of documents. Often, documents are part of a series or a conversation where the relevance to a query can be influenced by preceding or succeeding documents. By acknowledging this sequence, we can better capture contextual relationships that might be missed if documents were considered in isolation.

To formalize this process, we integrate GPT-3.5 Turbo's outputs into a mathematical model. This model takes into account the sequential dependencies among documents, treating the task of relevance detection as a problem of finding the optimal path or chain of relevance across the sequence.
%We observe that for Tamil and Kannada, GPT models have significant room for improvement.

\section{Related Work}

Code-mixing and transliteration have gained increasing attention in the field of natural language processing (NLP), especially as global communication becomes more digital and multilingual \cite{jauhiainen2019survey, muthusamy1994automatic, cm10}. This section reviews key studies related to code-mixing, information retrieval from code-mixed text, and the challenges of processing Roman transliterated languages, particularly in the context of Indian languages. Code-mixing, where speakers blend elements from multiple languages within a single utterance, is a common linguistic phenomenon in multilingual societies \cite{cm10}. Early studies on code-mixing focused primarily on sociolinguistic aspects, examining how and why speakers switch languages within conversations \cite{jauhiainen2019survey, muthusamy1994automatic, cm10}. However, with the advent of digital communication, researchers have increasingly turned their attention to computational methods for processing and understanding code-mixed text \cite{cm18}.

Several studies have explored various NLP tasks, such as part-of-speech tagging, language identification, and sentiment analysis, in code-mixed settings \cite{cm19}. \cite{cm11} provided one of the earliest comprehensive analyses of code-mixed text, highlighting the unique challenges it poses for traditional NLP pipelines, such as non-standard spelling, syntax variations, and the blending of multiple languages within a single text. More recent work by \cite{cm12} introduced a code-mixed dataset, spanning multiple Indian languages, which has become a benchmark for evaluating NLP models in this domain.

%Information Retrieval from Code-Mixed Text
Information retrieval (IR) in code-mixed settings is relatively underexplored compared to other NLP tasks \cite{cm17}. However, the need for effective IR systems that can handle multilingual and code-mixed queries has become increasingly important, particularly in the context of digital information exchange on social media platforms. \cite{cm13} investigated the problem of query-focused summarization in code-mixed social media data, emphasizing the complexity of extracting relevant information from noisy, informal text. Work by \cite{cm14} addressed code-mixed question answering, where the goal is to identify correct responses from a mixed-language corpus. Their approach involved using translation models to standardize the text before applying traditional IR techniques, demonstrating that even simple translation-based methods can significantly improve performance. However, these methods often fail to capture the nuances of code-mixed language, such as cultural context and colloquial expressions.

%Roman Transliteration and Indian Languages
Roman script transliteration of Indian languages, commonly referred to as "Romanagari"~\cite{mhaiskar2015romanagari} for languages like Hindi, is a widespread practice in digital communication. Transliteration introduces additional challenges for NLP, as it often involves non-standard spellings and inconsistent usage. For instance, multiple transliterations may exist for the same word, depending on the speaker's regional accent, literacy in the original script, or personal preference.

Notable efforts in this area include the work by \cite{cm15}, which explored transliteration normalization for Hindi-English code-mixed text. They developed algorithms to map Romanized text back to its original script, enabling more accurate processing by traditional NLP models. However, normalization remains a challenging task due to the inherent variability in transliterated text. In the context of Bengali, the Roman script transliteration is less standardized than for Hindi, leading to even greater variability in spelling and grammar. \cite{cm16} addressed this issue by creating a Roman Bengali dataset and proposed methods for transliteration normalization and language identification. Their work highlights the difficulties of processing Roman Bengali and the need for specialized approaches tailored to the characteristics of the language.

%Relevance to Our Work
While these studies provide valuable insights into code-mixing, transliteration, and information retrieval, there is a noticeable gap in addressing the specific challenges of extracting relevant information from code-mixed conversations in Roman transliterated Bengali. Our work builds on the foundations laid by previous research but focuses on the unique intersection of these challenges in a real-world context. By developing a mechanism to identify relevant answers in code-mixed discussions, we aim to contribute to the growing body of research on multilingual NLP and enhance the accessibility of information in linguistically diverse online communities.

Large Language Models (LLMs)~\cite{radford2019language,raffel2020exploring,liu2019roberta} like GPT-3 have shown promise in various NLP tasks, including LI. Previous works have demonstrated the capability of GPT-3 in performing zero-shot and few-shot learning, making it a potentially powerful tool for LI in resource-constrained settings. However, the application of LLMs~\cite{zhao2023survey,vaswani2017attention,gpt2-fine-tuning,zellers2019defending} to code-mixed and morphologically rich languages remains underexplored. Recent studies, have started to explore the use of transformers and pre-trained models for multilingual LI, but the effectiveness of these models in Bengali languages requires further investigation.

This section places our work within the context of existing research, highlighting the contributions of prior studies while identifying gaps that our research aims to fill.

\section{Dataset}

This shared task consists of a single dataset~\cite{ChandaP23} for code mixed information retrieval. The corpus consists of 107900 documents in the training set and 20 queries in the training set. There are 30 queries in the testing set. The dataset is in roman transliterated bengali mixed with english language.

%\end{enumerate}

\section{Task Definition}
%Objective:
The task~\footnote{\url{https://cmir-iitbhu.github.io/cmir/}} is to automatically determine the relevance of a query to a document within code-mixed data, specifically focusing on English and Roman transliterated Bengali.

Given a query and a document, the goal is to classify whether the query is relevant or not relevant to the document. Based on the relevance we have to rank the documents. This involves handling the complexities of code-mixing, where elements from both languages are used within the same text, and dealing with the informal and non-standardized nature of the language. The system must accurately capture the semantic relationship between the query and the document despite these linguistic challenges.

\section{Methodology}

\subsection{Why Prompting?}
Prompting \cite{cm25} for Information Retrieval is a burgeoning approach that leverages large language models (LLMs) to enhance the retrieval of relevant information from complex, unstructured data, such as code-mixed text or informal online conversations \cite{cm25}. Below are several reasons why prompting is becoming an effective strategy in information retrieval (IR):
\begin{itemize}[-]
    
\item \textbf{Handling Ambiguity and Contextual Nuances:}
Traditional IR systems often struggle with understanding the nuanced language, ambiguity, and context found in unstructured or informal text, such as code-mixed conversations. Prompting LLMs allows these models to interpret context more effectively by guiding them to generate or rank responses that are contextually appropriate, even when dealing with code-mixing or informal language structures \cite{cm20}. By crafting specific prompts, users can elicit more relevant and accurate results that account for the complexities of the input text.

\item \textbf{Enhanced Language Understanding:}
Large language models like GPT-3.5 are pre-trained on vast datasets that include a variety of languages and dialects \cite{cm21}. This extensive training enables them to understand and generate text across different languages and contexts \cite{cm21}. By using prompting, these models can be directed to focus on the most relevant aspects of a query or document, improving the retrieval process even in multilingual and code-mixed scenarios. For example, when retrieving information from Roman transliterated Bengali mixed with English, an LLM can be prompted to recognize and process the code-mixed language more effectively than traditional IR systems.

\item \textbf{Adaptability to Informal and Unstructured Text:}
Prompting allows LLMs to adapt to the informal and often unstructured nature of social media text \cite{zgheib2020social}, which is common in online communities. This flexibility is particularly beneficial when dealing with code-mixed or transliterated text, where the lack of standardization poses a challenge to conventional IR techniques. Prompted language models can generate or filter responses that align more closely with the informal tone and style of the original text, thereby improving the relevance of the retrieved information.

\item \textbf{Reduction of Noise and Irrelevance:}
One of the major challenges in IR is filtering out irrelevant or noisy data, especially in informal online conversations where off-topic or redundant information is common. By using targeted prompts, LLMs can be instructed to prioritize certain types of information, such as direct answers to specific questions, while de-emphasizing or ignoring irrelevant content \cite{cm23}. This leads to a more efficient and effective retrieval process, particularly in environments where users are seeking specific answers within a sea of mixed and informal language.

\item \textbf{Scalability and Customization:}
Prompting for information retrieval offers scalability and customization that traditional IR systems might lack. By designing prompts tailored to specific contexts or types of queries, LLMs can be dynamically adjusted to meet the needs of different retrieval tasks \cite{cm23}. This customization is particularly useful in handling domain-specific language or code-mixed scenarios, where standard IR systems might require extensive re-training or re-configuration.

\item \textbf{Real-Time Processing and Interaction:}
In real-time communication platforms, the ability to quickly retrieve relevant information based on ongoing conversations is crucial. Prompting enables LLMs to process and respond to queries in real-time, enhancing the interactivity and responsiveness of the IR system \cite{cm23}. This is especially beneficial in scenarios where users are engaged in active discussions and require immediate, contextually relevant information.

\end{itemize}

\subsection{Merging Prompt and Mathematical Model-Based Approaches}

We used the GPT-3.5 Turbo model via prompting through the OpenAI API\footnote{\url{https://platform.openai.com/docs/models/gpt-3-5-turbo}} to solve the document retrieval task. The process begins by first converting all the code-mixed sentences to english for both the queries and the documents. After this we try to determine the relevance scores, we used the following prompt:

"\textit{Given the query <query> and the document <document>, find how relevant is the query to the document based on semantic similarity. Provide a relevance score between 0 and 1. Only state the score}."

After the prompt is provided to the LLM, the following steps happen internal to the LLM while generating the output. The following outlines the steps that occur internally within the LLM, summarizing the prompting approach using GPT-3.5 Turbo:\\

\textbf{Step 1: Tokenization}

\begin{itemize}
    \item \textbf{Prompt:} \( X = [x_1, x_2, \dots, x_n] \)
    \item The input text (prompt) is first tokenized into smaller units called tokens. These tokens are often subwords or characters, depending on the model's design.
    \item \textbf{Tokenized Input:} \( T = [t_1, t_2, \dots, t_m] \)
\end{itemize}

\textbf{Step 2: Embedding}

\begin{itemize}
    \item Each token is converted into a high-dimensional vector (embedding) using an embedding matrix \( E \).
    \item \textbf{Embedding Matrix:} \( E \in \mathbb{R}^{|V| \times d} \), where \( |V| \) is the size of the vocabulary and \( d \) is the embedding dimension.
    \item \textbf{Embedded Tokens:} \( T_{\text{emb}} = [E(t_1), E(t_2), \dots, E(t_m)] \)
\end{itemize}

\textbf{Step 3: Positional Encoding}

\begin{itemize}
    \item Since the model processes sequences, it adds positional information to the embeddings to capture the order of tokens.
    \item \textbf{Positional Encoding:} \( P(t_i) \)
    \item \textbf{Input to the Model:} \( Z = T_{\text{emb}} + P \)
\end{itemize}

\textbf{Step 4: Attention Mechanism (Transformer Architecture)}

\begin{itemize}
    \item \textbf{Attention Score Calculation:} The model computes attention scores to determine the importance of each token relative to others in the sequence.
    \item \textbf{Attention Formula:}
    \begin{equation}
    \text{Attention}(Q, K, V) = \text{softmax}\left(\frac{QK^T}{\sqrt{d_k}}\right)V
    \end{equation}
    \item where \( Q \) (query), \( K \) (key), and \( V \) (value) are linear transformations of the input \( Z \).
    \item This attention mechanism is applied multiple times through multi-head attention, allowing the model to focus on different parts of the sequence simultaneously.
\end{itemize}

\textbf{Step 5: Feedforward Neural Networks}

\begin{itemize}
    \item The output of the attention mechanism is passed through feedforward neural networks, which apply non-linear transformations.
    \item \textbf{Feedforward Layer:}
    \begin{equation}
    \text{FFN}(x) = \max(0, xW_1 + b_1)W_2 + b_2
    \end{equation}
    \item where \( W_1, W_2 \) are weight matrices and \( b_1, b_2 \) are biases.
\end{itemize}

\textbf{Step 6: Stacking Layers}

\begin{itemize}
    \item Multiple layers of attention and feedforward networks are stacked, each with its own set of parameters. This forms the "deep" in deep learning.
    \item \textbf{Layer Output:}
    \begin{equation}
    H^{(l)} = \text{LayerNorm}(Z^{(l)} + \text{Attention}(Q^{(l)}, K^{(l)}, V^{(l)}))
    \end{equation}
    \begin{equation}
    Z^{(l+1)} = \text{LayerNorm}(H^{(l)} + \text{FFN}(H^{(l)}))
    \end{equation}
\end{itemize}

\textbf{Step 7: Output Generation}

\begin{itemize}
    \item The final output of the stacked layers is a sequence of vectors.
    \item These vectors are projected back into the token space using a softmax layer to predict the next token or word in the sequence.
    \item \textbf{Softmax Function:}
    \begin{equation}
    P(y_i|X) = \frac{\exp(Z_i)}{\sum_{j=1}^{|V|} \exp(Z_j)}
    \end{equation}
    \item where \( Z_i \) is the logit corresponding to token \( i \) in the vocabulary.
    \item The model generates the next token in the sequence based on the probability distribution, and the process repeats until the end of the output sequence is reached.
\end{itemize}

\textbf{Step 8: Decoding}

\begin{itemize}
    \item The predicted tokens are then decoded back into text, forming the final output.
    \item \textbf{Output Text:} \( Y = [y_1, y_2, \dots, y_k] \)
\end{itemize}

%After getting the relevance score we used the following mathematical formulation to allow for sequential presence of relevant documents,

%if the score of the current document %\(D_n\) is less than 0.1, and you want the probability for the next document to be just the score itself, the formulation can be updated with a conditional expression.
After obtaining the relevance score, we used the following mathematical formulation to account for the sequential presence of relevant documents. This can be written as follows:

\[
P(D_{n+1} \mid D_n) =
\begin{cases} 
\text{Score}(D_{n+1}) & \text{if } \text{Score}(D_{n+1}) < 0.3 , D_n=relevant\\
\text{Score}(D_{n+1}) & \text{if } n = -1 \\
0.2 + \text{Score}(D_{n+1}) & \text{if } \text{Score}(D_{n+1}) >= 0.3 , D_n=relevant\\ 
\text{Score}(D_{n+1}) & otherwise
\end{cases}
\]

This equation now reflects that if the score of the current document \(D_n\) is less than 0.3 and the previous document is relevant, the probability of the current document being relevant is simply equal to the relevance score of current document.

If the previous document is relevant and if the score of the current document \(D_n\) is greater than equal to 0.3 then the probability that the current document is relevant is 0.2 + Score for the current document.
For the first document, the probability is equal to the relevance score of current document. In all other situations, the probability is equal to the relevance score of current document. If the probability score of a particular document is greater than 0.5, we consider the document to be relevant to the query. Like this we found out all documents which are relevant to a query.

An example of the mathematical formulation and how it helps to detect relevant documents is shown in Table~\ref{tab:query-document}. The table reflects a range of documents to a code mixed query. The relevance scores help identify how closely each document addresses the query, while the probability scores provide insight into the potential usefulness of the documents based on the provided content. Overall, Documents 1-4 stands out as particularly relevant based on probability scores.

%\(\text{Score

\begin{figure}[h!]
  \centering
  \includegraphics[width=0.54\linewidth]{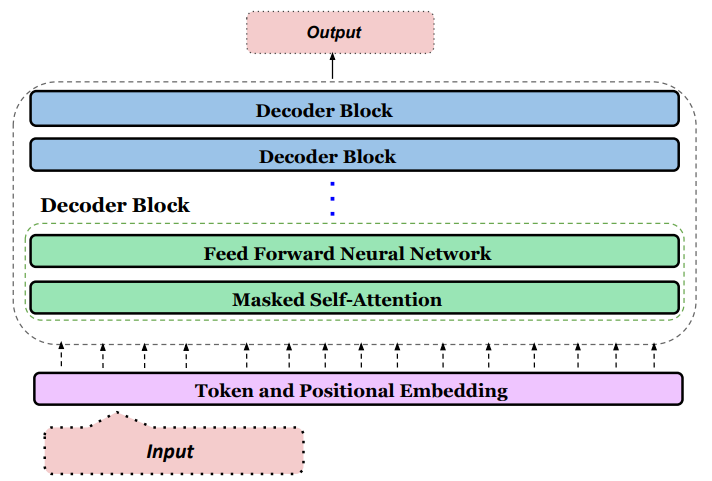}
  \caption{An overview of the GPT-3.5 Turbo architecture.} \label{fig1}
  %\Description{A woman and a girl in white dresses sit in an open car.}
\end{figure}

\begin{figure}[h!]
  \centering
  \includegraphics[width=0.54\linewidth]{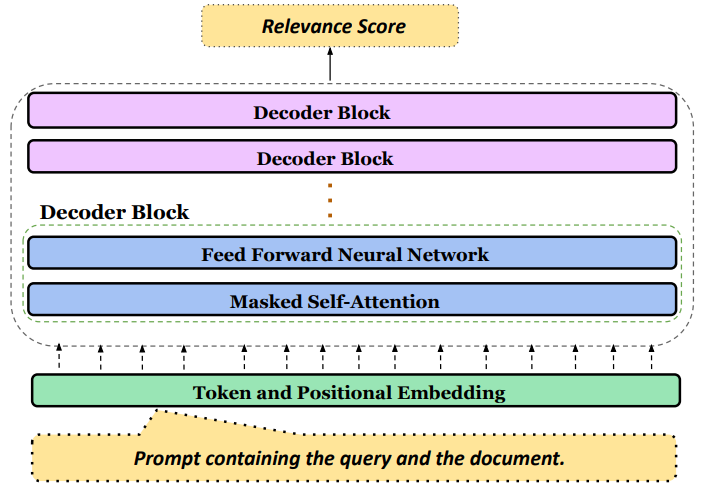}
  \caption{Overview diagram of the methodology followed for GPT-3.5 Turbo.} \label{fig2}
  %\Description{A woman and a girl in white dresses sit in an open car.}
\end{figure}

For the five results reported, we ran the GPT model at different temperature values namely 0.5, 0,6, 0.7, 0.8, and 0.9. The diagram for GPT-3.5 Turbo is shown in Figure~\ref{fig1}. The figure representing the methodology is shown in Figure~\ref{fig2}.

At lower temperatures, the model's responses are more deterministic and focused. It generates outputs that are likely to be relevant and closely aligned with the input, making it useful for tasks requiring precision, such as retrieving specific information or handling queries with clear intent. 
At higher temperatures results in highly diverse and less predictable outputs. It can be useful in exploratory tasks where creativity and variation are needed, but it may also risk generating less coherent or relevant responses. 
In code-mixed scenarios, this could capture the full spectrum of linguistic creativity but might require careful handling to ensure relevance.
So we used a temperature range of 0.5 to 0.9.

\begin{table}[h]
    \centering
    \begin{tabular}{@{}>{\raggedright}p{3cm} >{\raggedright}p{6cm} l l@{}}
        \toprule
        \textbf{Query} & \textbf{Document} & \textbf{Relevance Score} & \textbf{Probability Score} \\ \midrule
        
        Kivabe bhalo bhabe ingreji shikhbo? & Shudhumatro lekhar opor focus korle hobe na. \\ 
        Ingreji songs shunle shikha sohoj hoy. & 0.55 & 0.55 (1) \\
\hline
        Kivabe bhalo bhabe ingreji shikhbo? & Ingreji film dekhle vocabulary bere jaye. \\ 
        Conversation practice korao khub helpful. & 0.45 & 0.65 (1) \\
\hline
        Kivabe bhalo bhabe ingreji shikhbo? & Ingreji shikhar jonyo grammar book khub important. \\ 
        Kintu speaking practice beshi dorkar. & 0.35 & 0.55 (1) \\
\hline
        Kivabe bhalo bhabe ingreji shikhbo? & Bhasha shikhte bhalo tutor nirbachon kora joruri. \\ 
        Bibhinno apps byabohar kore practice kora jaye. & 0.45 & 0.65 (1) \\
\hline
        Kivabe bhalo bhabe ingreji shikhbo? & time lagbe. & 0.20 & 0.20 (0) \\ \bottomrule
    \end{tabular}
    \caption{Example of Query and Document Relevance along with probability scores and relevance scores. Beside the probability score the relevance of a document to a query is stated in braces. 1 represents relevant and 0 represents not-relevant.}
    \label{tab:query-document}
\end{table}

%\noindent \textbf{(a)} We used the following prompt for Kannada language for the purpose of classification: 

%"\textit{Please identify which category the word is in English, Kannada, Mixed, Name, Location, Symbol and Other. Please state En, Kn, Mixed, Name, Location, sym and Other. The word is <Word>}."

%The figure representing the methodology is shown in Figure ~\ref{fig2}.

%For the five results reported, we ran the GPT model at different temperature values ranging from: {0.5, 0,6, 0.7, 0.8, and 0.9}.

%\noindent \textbf{(b)} We used the following prompt for Tamil language for the purpose of classification: 

%"\textit{Please identify which category the word is in English, Tamil, Mixed, Name, Location, Symbol and Other. Please state en, tm, tmen, name, Location, sym and Other. The word is <Word>.}"

%The figure representing the methodology is shown in Figure ~\ref{fig3}.
\section{Results}

\begin{table}[h!]
\centering
\begin{tabular}{c|c|c|c|c|c|c}
\toprule
\toprule
\textbf{MAP Score} & \textbf{ndcg Score} & \textbf{p@5 Score} & \textbf{p@10 Score} & \textbf{Team Name} & \textbf{Submission File} & Rank\\ \midrule \midrule
0.701773  & 0.797937   & 0.793333  & 0.766667   & TextTitans  & submit\_cmir & 5        \\ \midrule
0.701773  & 0.797937   & 0.793333  & 0.766667   & TextTitans  & submit\_cmir\_1 & 4      \\ \midrule
0.701773  & 0.797937   & 0.793333  & 0.766667   & TextTitans  & submit\_cmir\_2 & 3      \\ \midrule
0.701773  & 0.797937   & 0.793333  & 0.766667   & TextTitans  & submit\_cmir\_3 & 2      \\ \midrule
0.703734  & 0.799196   & 0.793333  & 0.766667   & TextTitans  & submit\_cmir\_4 & 1      \\ \bottomrule \bottomrule
\end{tabular}
\caption{A Comparison of MAP, NDCG, P@5, and P@10 Scores for the TextTitans Team.}
\label{tab:score_comparison}
\end{table}

Table~\ref{tab:score_comparison} presents the evaluation metrics for different submissions~\cite{chanda2024overview} for the team named "TextTitans". The metrics used to assess the performance are MAP Score, ndcg Score, p@5 Score, and p@10 Score. Here's what these results imply. MAP is a common metric in information retrieval that measures the precision of results across multiple queries. A higher MAP score indicates that relevant documents are consistently ranked higher across all queries. In the table, the MAP scores for the first four submissions are identical (0.701773), while the fifth submission slightly improves to 0.703734. This indicates that the fifth submission is marginally better in terms of ranking relevant results across multiple queries.
The ndcg score measures the quality of the ranking based on the position of relevant documents. A higher ndcg score suggests that relevant documents are placed higher in the ranking. The scores are also very similar across submissions, with the first four submissions having an ndcg score of 0.797937, and the fifth submission showing a slight improvement to 0.799196. This suggests a minor improvement in ranking relevant documents for the fifth submission. p@5 measures how many of the top 5 ranked documents are relevant. A score of 1 would mean that all 5 of the top-ranked documents are relevant. All submissions have the same p@5 score of 0.793333, indicating that the top 5 results are equally accurate across all submissions.
Precision@10 measures how many of the top 10 ranked documents are relevant. Like p@5, a higher score is better. Similar to p@5, all submissions have the same p@10 score of 0.766667, showing no variation in the top 10 results across the different submissions. The metrics are very consistent across all submissions, with only minor improvements in MAP and NDCG scores for the fifth submission. The fifth submission shows a slight improvement in ranking and retrieval performance, but the changes are minimal. The p@5 and p@10 scores indicate that the precision of the top 5 and top 10 results is identical across all submissions, suggesting that the models are performing similarly in identifying the most relevant documents. Overall, while there is a slight improvement in the last submission, the models generally perform similarly across all metrics.

\section{Conclusion}
%Conclusion
In conclusion, this study addresses the critical challenges of extracting relevant information from code-mixed conversations, specifically within Roman transliterated Bengali mixed with English. This linguistic phenomenon is prevalent among migrant communities in India, who often rely on social media platforms to share and seek vital information, especially during crises like the COVID-19 pandemic. The informal and non-standardized nature of these conversations presents unique difficulties for information retrieval. To tackle these challenges, we developed a novel approach that leverages the GPT-3.5 Turbo model in conjunction with a sequential engineering approach, achieving notable success in retrieving pertinent answers from complex, code-mixed digital conversations. The effectiveness of our method is demonstrated through the results on the test set documents and queries, which provides a valuable resource for future research in natural language processing within multilingual and informal text environments. This work contributes to enhancing information accessibility for marginalized communities, underscoring the potential of advanced AI models in bridging communication gaps in diverse linguistic landscapes. We observe that the GPT-3.5 model along with mathematical formulation approach performs well for the task of Code mixed information retrieval, though there is scope for improvement.

%\begin{acknowledgments}
%  Thanks to the developers of ACM consolidated LaTeX styles
%  \url{https://github.com/borisveytsman/acmart} and to the developers
%  of Elsevier updated \LaTeX{} templates
%  \url{https://www.ctan.org/tex-archive/macros/latex/contrib/els-cas-templates}.  
%\end{acknowledgments}

%%
%% Define the bibliography file to be used
\bibliography{sample-1col}

\appendix


%The sources for the ceur-art style are available via
%\begin{itemize}
%\item \href{https://github.com/yamadharma/ceurart}{GitHub},
% \item \href{https://www.overleaf.com/project/5e76702c4acae70001d3bc87}{Overleaf},
%\item
%  \href{https://www.overleaf.com/latex/templates/template-for-submissions-to-ceur-workshop-proceedings-ceur-ws-dot-org/pkfscdkgkhcq}{Overleaf
%    template}.
%\end{itemize}

\end{document}